\documentclass[11pt]{article}
\usepackage[utf8]{inputenc}
\usepackage[T1]{fontenc}
\usepackage{lmodern}
\usepackage{microtype}

\usepackage[margin=1in]{geometry}
\usepackage{amsmath, amssymb}

\usepackage{graphicx}
\usepackage{caption}
\usepackage{subcaption}
\usepackage{float}      
\usepackage{placeins}   
\graphicspath{{samples/}}  

\usepackage{algorithm}
\usepackage{algpseudocode} 

\usepackage[numbers]{natbib}
\usepackage{booktabs}
\usepackage{hyperref}

\title{QSilk: Micrograin Stabilization and Adaptive Quantile Clipping for Detail-Friendly Latent Diffusion}
\author{
Denis Rychkovskiy ("DZRobo", Independent Researcher)\\
\texttt{nebularus@yandex.ru}
}
\date{October 17, 2025}

\begin{document}
\maketitle
\vspace{-0.5em}
\begin{center}
{\large Primary Subject Class: cs.CV \quad Secondary Class: cs.LG}
\end{center}

\begin{abstract}
We present \textbf{QSilk}, a lightweight, always-on stabilization layer for latent diffusion that improves high-frequency fidelity while suppressing rare activation spikes. QSilk combines (i) a per-sample micro-clamp that gently limits extreme values without washing out texture, and (ii) \emph{Adaptive Quantile Clip} (AQClip), which adapts the allowed value corridor per region. AQClip can operate in a proxy mode using local structure statistics or in an Attn mode guided by attention entropy (model confidence). Integrated into the CADE~2.5 rendering pipeline, QSilk yields cleaner, sharper results at low step counts and ultra-high resolutions with negligible overhead. It requires no training, no model finetuning, and exposes minimal user controls. We report consistent improvements across SD/SDXL backbones and show that QSilk synergizes with CFG/Rescale, enabling slightly higher guidance without artifacts.
\end{abstract}

\section{Introduction}
Large-scale diffusion models can exhibit unstable activation tails that manifest as halos, moir\'e, or ``grid'' artifacts---especially at high resolution, aggressive CFG, and low step counts. Na\"{\i}ve global clipping removes spikes but often dulls micro-texture. We seek a practical, training-free stabilizer that preserves detail while gently taming extremes and adapts to local confidence.

\paragraph{Contributions.} Our main contributions are:
\begin{itemize}
\item \textbf{QSilk micrograin stabilizer}: a gentle per-sample soft clamp that suppresses extreme latent values without flattening texture.
\item \textbf{AQClip (adaptive quantile clipping)}: per-tile, seam-free soft clipping whose corridor widens in confident regions and narrows where the model is uncertain.
\item \textbf{Attn-mode}: an attention-entropy--guided confidence map for AQClip that further refines the detail/cleanliness trade-off.
\item \textbf{Plug-and-play integration}: a minimal-overhead module for CADE~2.5 (SD/SDXL), with robust defaults and reproducible presets.
\end{itemize}

\section{Background}
\paragraph{Latent diffusion and CFG/Rescale.} We build on DDPM~\cite{ho2020ddpm} and latent diffusion~\cite{rombach2022ldm}. High CFG often amplifies activation tails; rescaled variants mitigate this trade-off.

\paragraph{Dynamic thresholding.} Imagen introduced dynamic thresholding in \emph{image space} to enable higher guidance without saturation~\cite{saharia2022imagen, saharia2022imagenSup}. Diffusers warns it is unsuitable in \emph{latent} space~\cite{diffusersDynamicThresh}. Our method differs by applying \emph{tile-wise quantile corridors in latent space} and by adapting the corridor to model confidence.

\paragraph{Attention and confidence.} Peaky, low-entropy attention often indicates local certainty, while diffuse, high-entropy attention signals uncertainty~\cite{pardyl2023attentionEntropy,huang2025vdmAttentionEntropy}. We leverage this via an entropy map captured at sampling-time.

\paragraph{Spatially varying guidance.} Spatially adaptive CFG (e.g., S-CFG) adjusts guidance per semantic region~\cite{shen2024scfg}. QSilk is complementary: it does not alter guidance; instead it regularizes latent amplitudes locally, reducing artifact tails that guidance may amplify.

\section{Method}
\subsection{QSilk Micrograin Stabilizer (global, per-sample)}
Given a denoised latent $x \in \mathbb{R}^{B\times C\times H\times W}$, compute per-sample low/high quantiles $(q_{\ell}, q_{h})$ (e.g., $0.1\%/99.9\%$), then apply a \emph{soft} clamp between them. A tanh form preserves contrast:
\begin{equation}
\begin{aligned}
\ell &= \mathrm{quantile}(x, q_{\ell}),\quad h = \mathrm{quantile}(x, q_{h}), \\
m &= \frac{\ell + h}{2},\ \ \ \delta = \frac{h-\ell}{2},\\
x' &= m + \delta \cdot \tanh\!\left(\alpha \cdot \frac{x-m}{\delta+\varepsilon}\right).
\end{aligned}
\end{equation}
In CADE~2.5 we use a fast hard-clamp variant by default for minimal overhead; the tanh form is available and yields similar behavior at slightly higher cost.

\subsection{AQClip-Lite (proxy confidence, per-tile, seam-free)}
We adapt the clip corridor per spatial tile using a \emph{proxy confidence} derived from the local gradient magnitude of the channel-mean latent. Let $T$ denote tile size and $S$ stride. On the pooled grid we compute a normalized confidence $\hat{H}\in[0,1]$ and map it to asymmetric quantiles:
\begin{equation}
q_{\ell} = 0.5\cdot \hat{H}^2,\quad q_{h} = 1 - 0.5 \cdot (1-\hat{H})^2.
\end{equation}
Assuming per-tile normality, we estimate $(\ell, h)$ from $(\mu,\sigma)$ via the Normal inverse $\mathrm{ndtri}$. We then apply a \emph{tanh soft-clip} in the unfold--fold domain with overlap-add normalization, and use EMA over $(\ell,h)$ across steps to avoid flicker.

\begin{algorithm}[t]
\caption{AQClip-Lite (one denoising step)}\label{alg:aqclip}
\begin{algorithmic}[1]
\State \textbf{Input:} latent $z\in\mathbb{R}^{B\times C\times H\times W}$, tile $T$, stride $S$, softness $\alpha$, EMA $\beta$
\State $z_m \gets \mathrm{mean\_channel}(z)$; $g \gets \mathrm{avgpool}(\|\nabla z_m\|, T,S)$; $\hat{H}\gets g/\max(g)$
\State $q_\ell \gets 0.5 \hat{H}^2$;\ \ $q_h \gets 1-0.5(1-\hat{H})^2$
\State $U \gets \mathrm{unfold}(z, T,S)$; $\mu\gets \mathrm{mean}(U)$; $\sigma\gets \mathrm{std}(U)$
\State $\ell\gets \mu - \mathrm{ndtri}(q_\ell)\sigma$;\ \ $h\gets \mu + \mathrm{ndtri}(q_h)\sigma$
\State $(\ell,h) \gets \mathrm{EMA}((\ell,h);\beta)$ \Comment{per-tile EMA across steps}
\State $y\gets \tanh\!\big(\alpha \cdot \frac{U - ( \ell+h)/2}{(h-\ell)/2 + \varepsilon}\big)$
\State $U' \gets ( \ell+h)/2 + (h-\ell) y/2$
\State $z' \gets \mathrm{fold}(U') / \mathrm{fold}(\mathbf{1})$
\State \textbf{return} $z'$
\end{algorithmic}
\end{algorithm}

\subsection{AQClip-Attn (attention-entropy confidence)}
Instead of proxy gradients, we use an \emph{attention-entropy} probe: subsample a few heads and tokens, compute $p=\mathrm{softmax}(QK^\top/\sqrt{d})$ and its per-query entropy; reshape to a grid and normalize to $[0,1]$ to obtain $\hat{H}$. Quantile mapping and soft-clip follow as in AQClip-Lite.

\subsection{Placement in the pipeline}
In CADE~2.5 we place QSilk/AQClip (i) after each sampling iteration (post-CFG), before VAE decode and any late HF polish; and (ii) before each decode/encode cycle in multi-pass workflows. This positioning preserves texture while preventing artifact growth.

\section{Integration with CADE~2.5}
\textbf{Components.} ZeResFDG (hybrid CFGZero/RescaleFDG)~\cite{rychkovskiy2025cade25zeresfdg}, NAG (normalized attention guidance), ControlFusion masks, EPS scale, Muse Blend, Polish. QSilk is also exposed in \texttt{sagpu\_attention} paths to stabilize attention-driven detail at sampling time. Our reference implementation provides robust defaults and toggles.

\textbf{Synergy.} AQClip reduces artifact tails, allowing $+0.5\sim+1.0$ higher effective CFG without speckle; ZeResFDG then sharpens detail safely.

\textbf{Defaults.} QSilk (global): on, $q_\ell{=}0.001$, $q_h{=}0.999$, $\alpha{\approx}2.0$. AQClip-Lite: $T{=}32$, $S{=}16$, $\alpha{=}2.0$, EMA $\beta{\approx}0.8$; applied after sampling and before decode (toggle). AQClip-Attn: same as Lite when the attention probe is enabled.

\section{Experiments}
\begin{figure}[tbp]
    \centering
    \begin{subfigure}{\textwidth}
        \centering
        \includegraphics[width=0.49\linewidth]{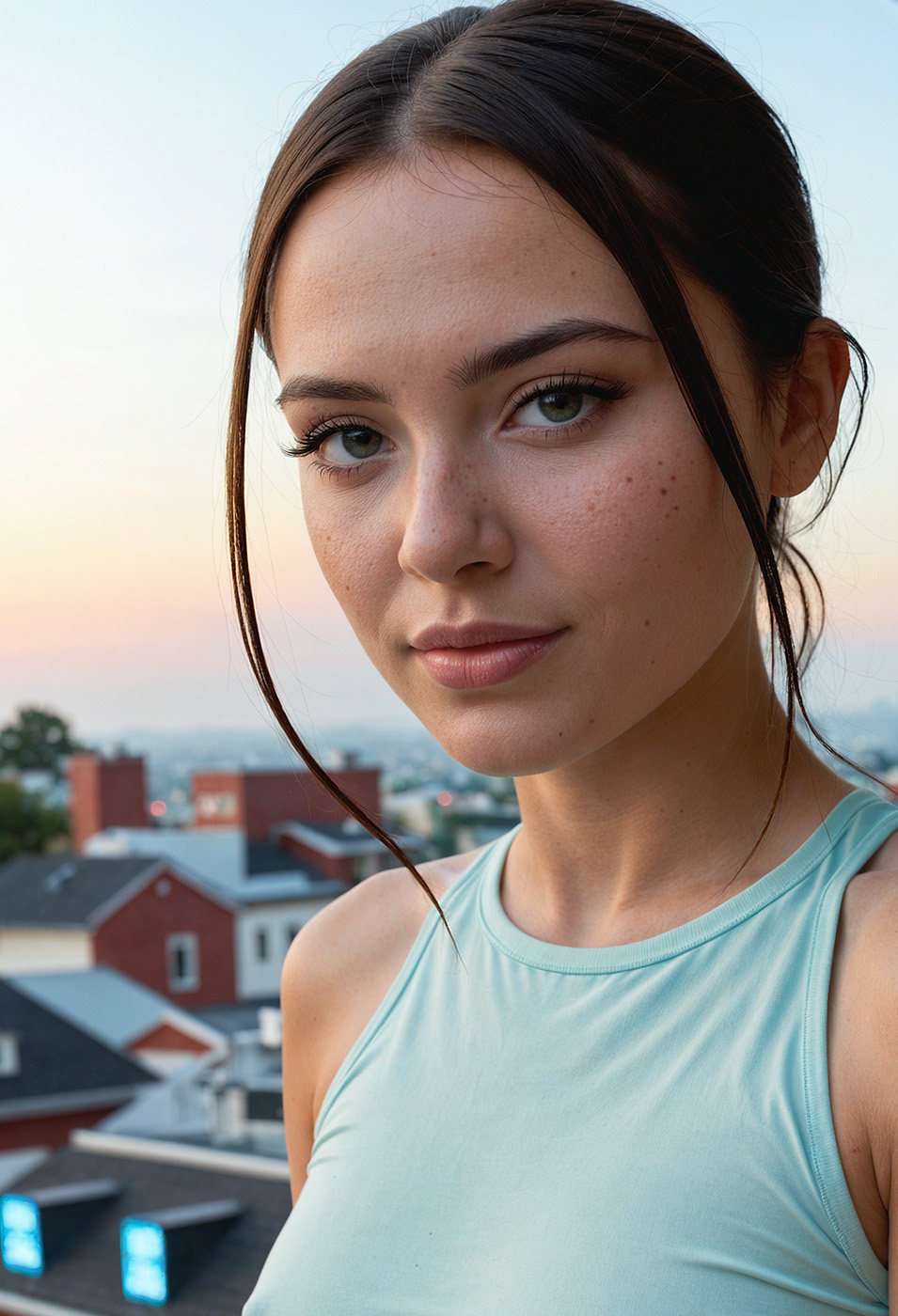}\hfill
        \includegraphics[width=0.49\linewidth]{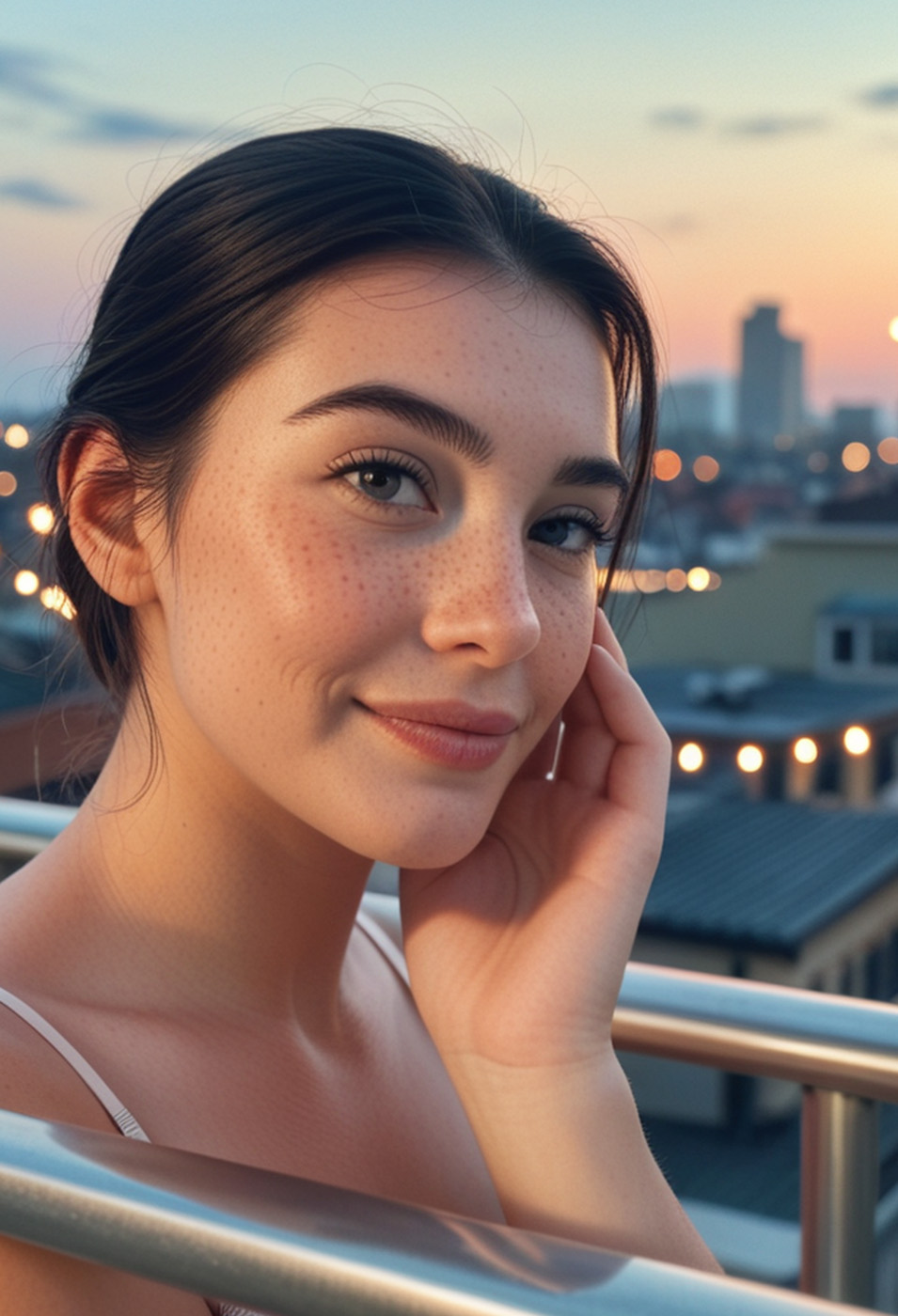}
        \caption{Photoportrait. Left: \textbf{CADE~2.5 + QSilk}. Right: baseline (KSampler).}
    \end{subfigure}
    \\
    \begin{subfigure}{\textwidth}
        \centering
        \includegraphics[width=0.49\linewidth]{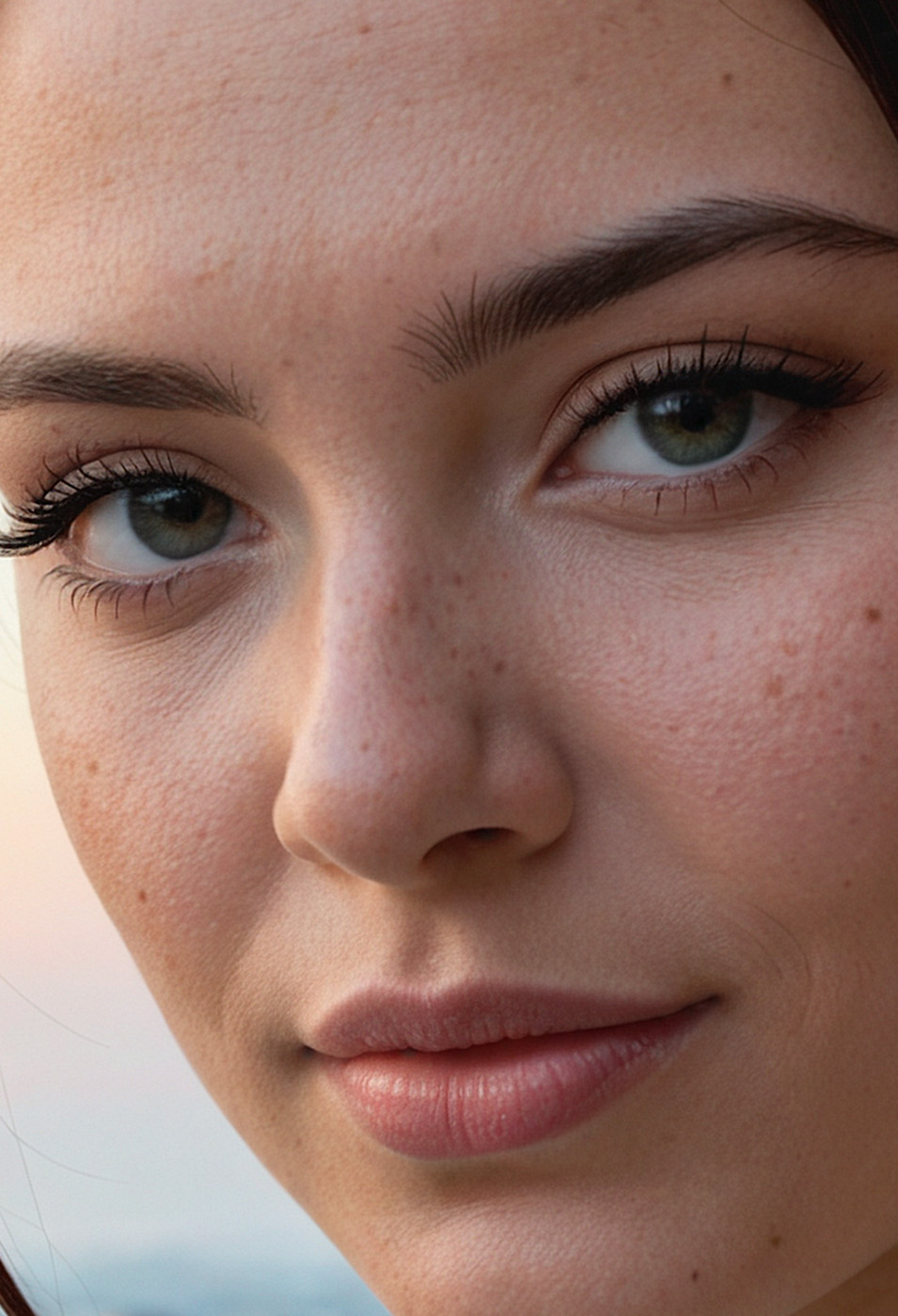}\hfill
        \includegraphics[width=0.49\linewidth]{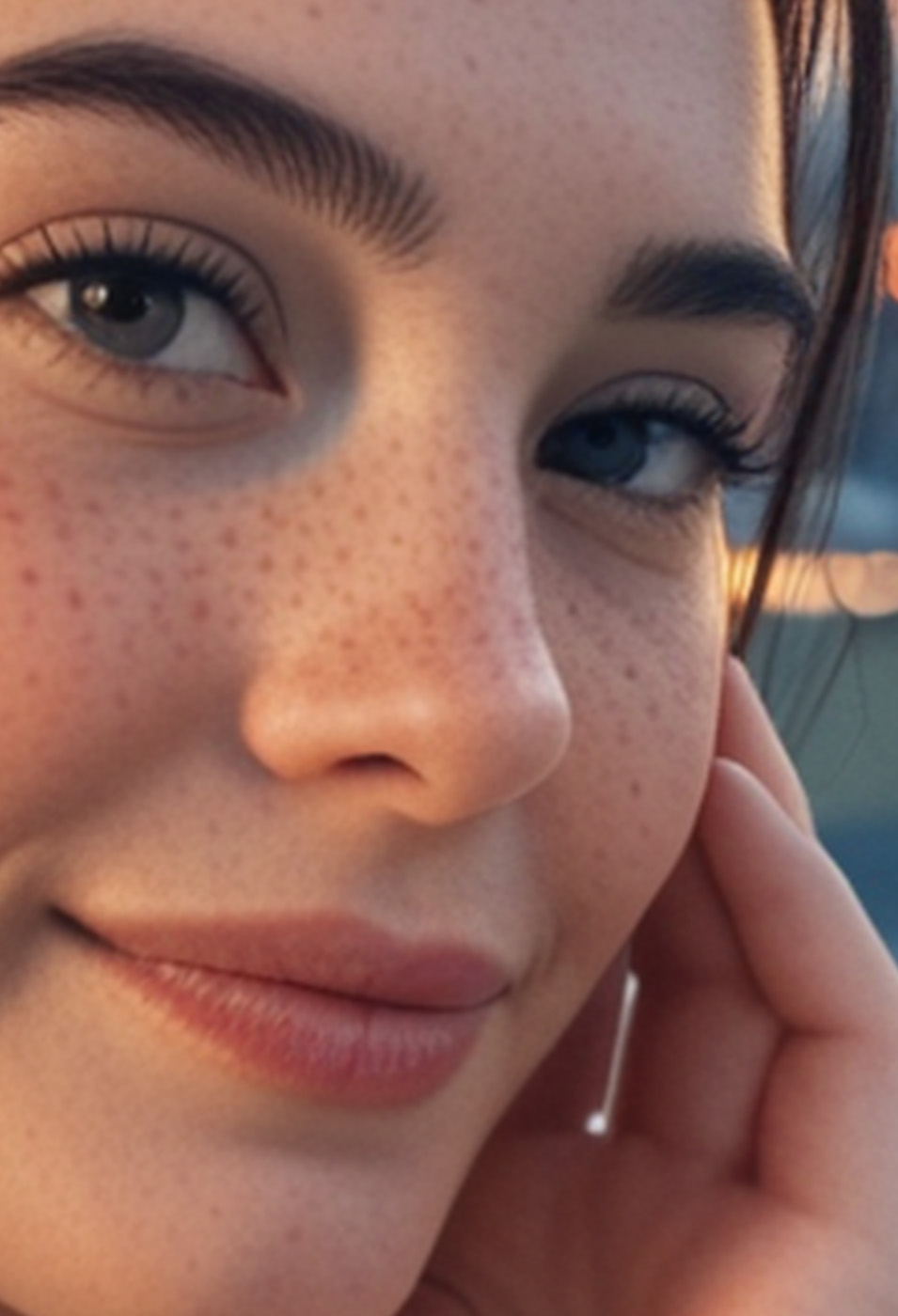}
        \caption{Photoportrait, zoom-in: coherent skin micrograin, cleaner HF detail.}
    \end{subfigure}
    \caption{Qualitative comparison (portrait). Left: CADE~2.5 with QSilk. Right: baseline without QSilk/AQClip. Same prompts and seed.}
    \label{fig:portrait_qual}
\end{figure}

\begin{figure}[tbp]
    \centering
    \begin{subfigure}{\textwidth}
        \centering
        \includegraphics[width=0.49\linewidth]{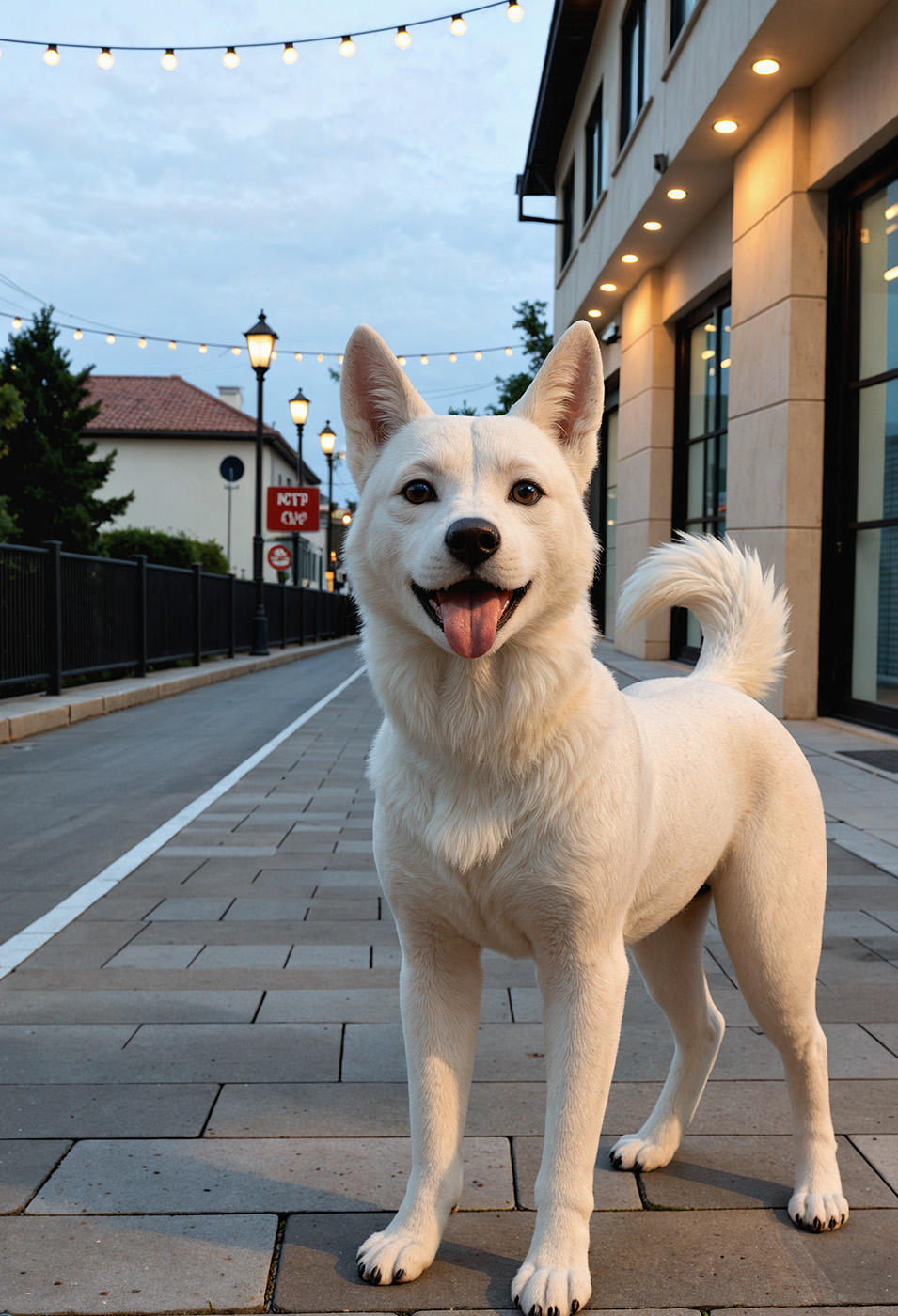}\hfill
        \includegraphics[width=0.49\linewidth]{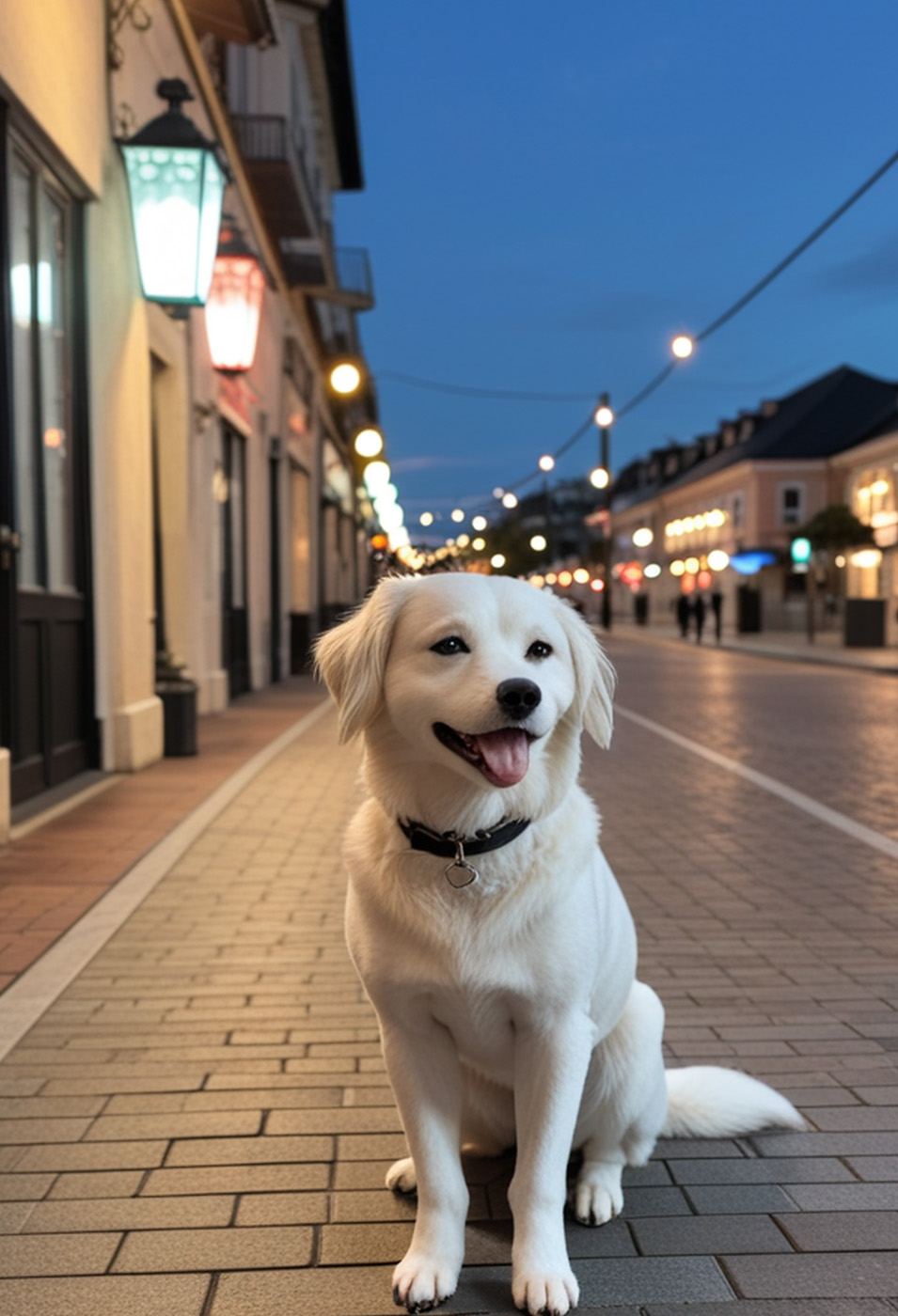}
        \caption{Dog. Left: \textbf{CADE~2.5 + QSilk}. Right: baseline.}
    \end{subfigure}
    \\
    \begin{subfigure}{\textwidth}
        \centering
        \includegraphics[width=0.49\linewidth]{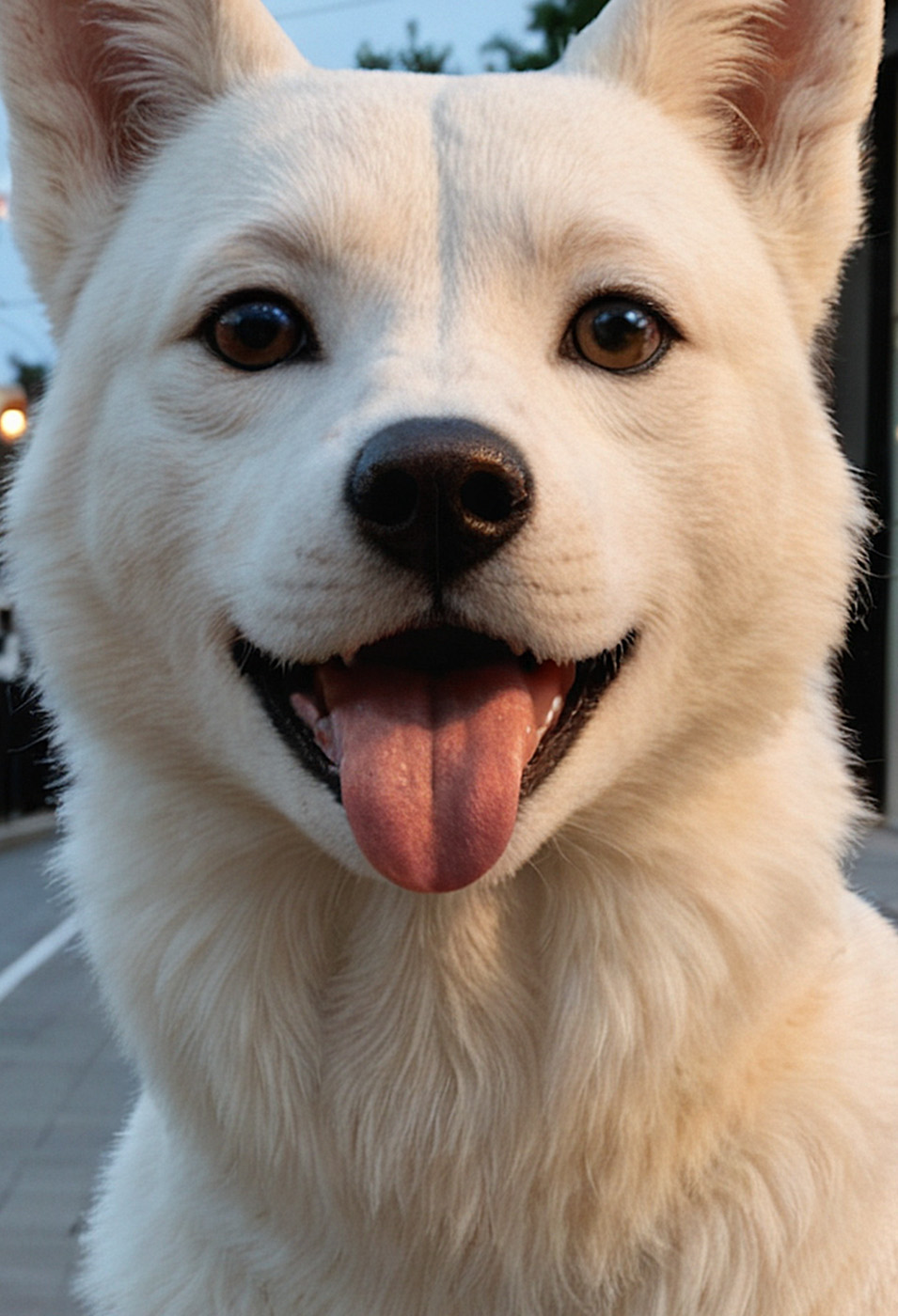}\hfill
        \includegraphics[width=0.49\linewidth]{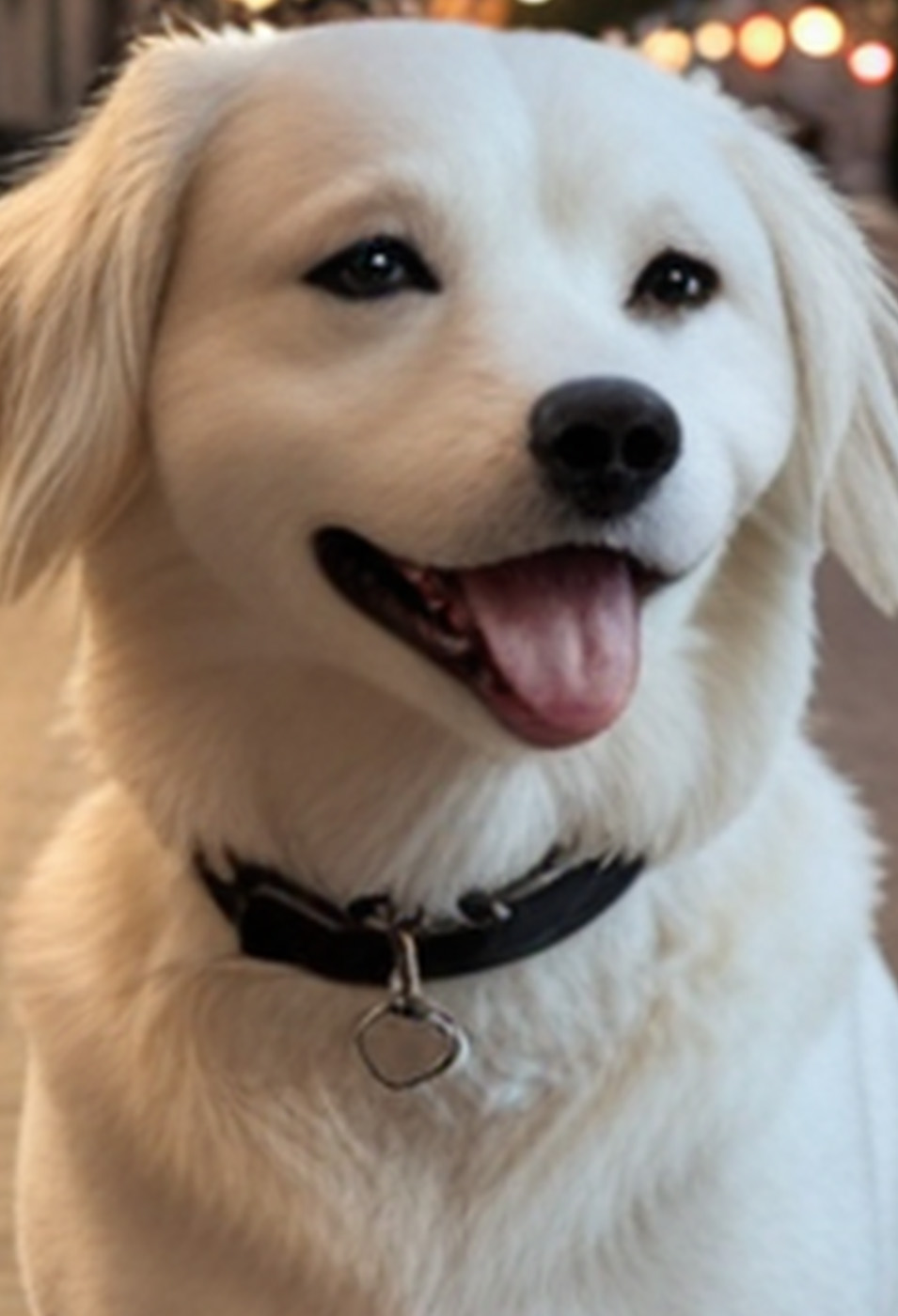}
        \caption{Dog, zoom-in: improved fur coherence, suppressed halo/moire.}
    \end{subfigure}
    \caption{Qualitative comparison (dog). Left: CADE~2.5 with QSilk. Right: baseline. Same prompts and seed.}
    \label{fig:dog_qual}
\end{figure}

\begin{figure}[tbp]
    \centering
    \begin{subfigure}{\textwidth}
        \centering
        \includegraphics[width=0.49\linewidth]{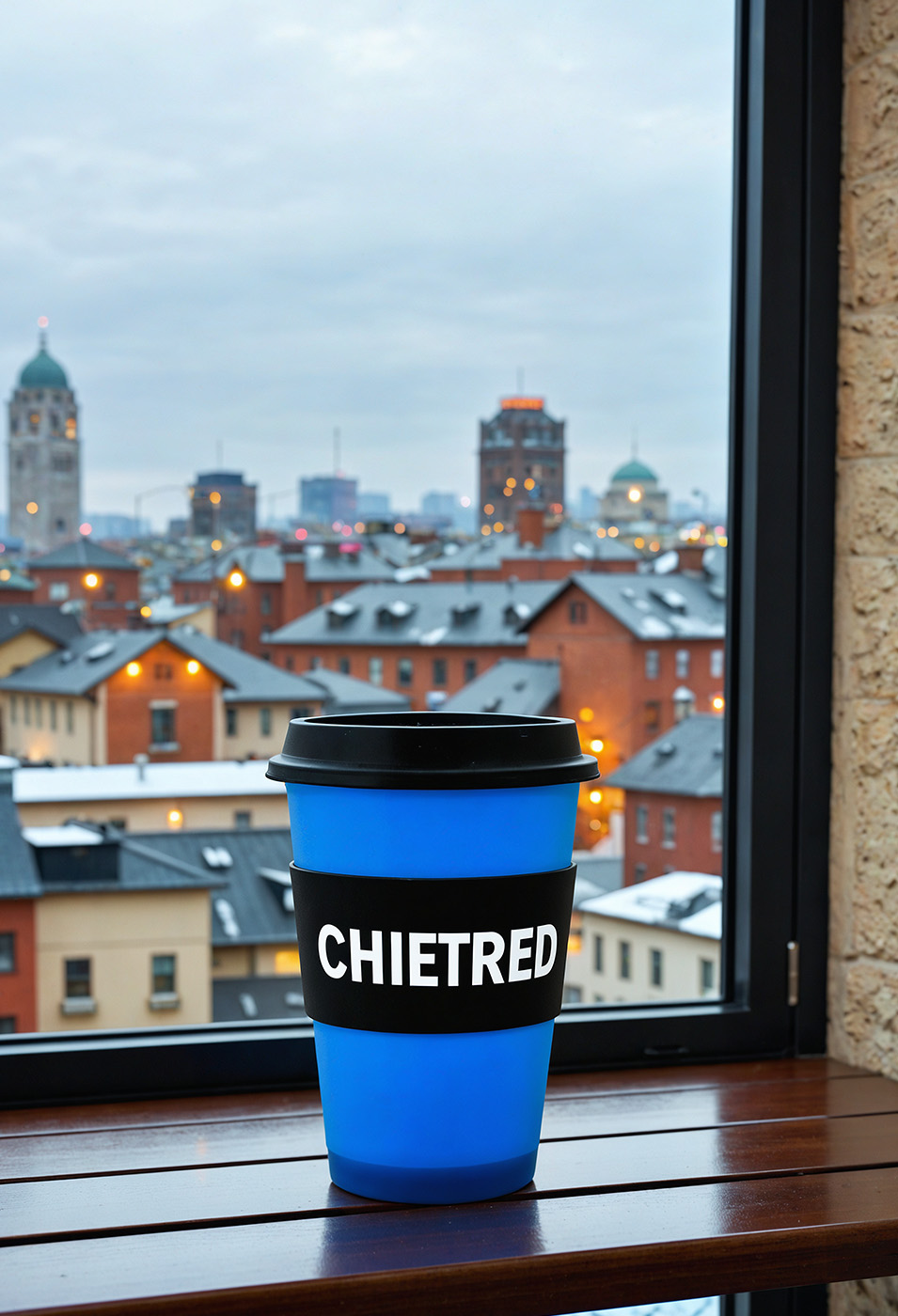}\hfill
        \includegraphics[width=0.49\linewidth]{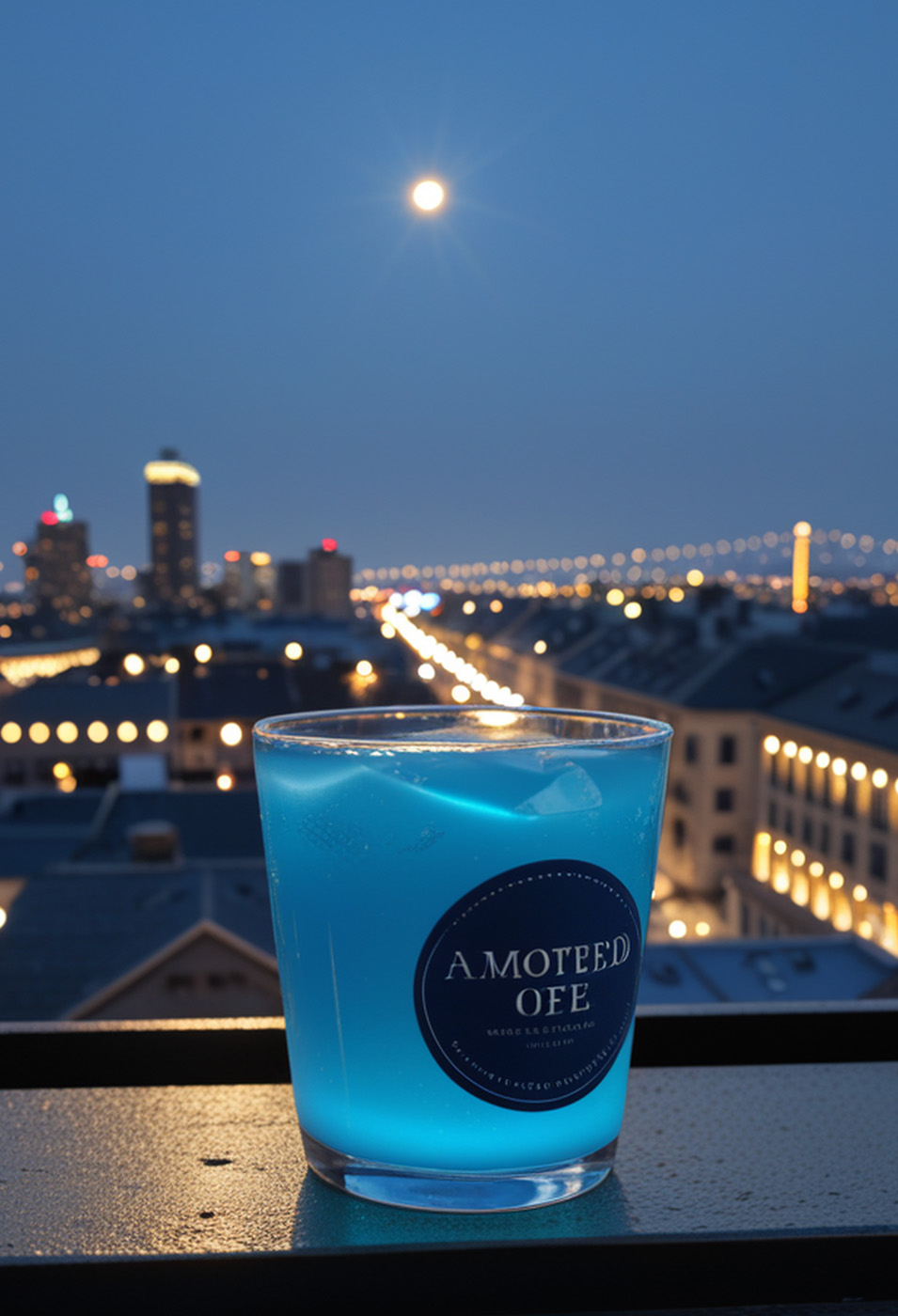}
        \caption{Cup. Left: \textbf{CADE~2.5 + QSilk}. Right: baseline.}
        \label{fig:cup_qual_main}
    \end{subfigure}
    \\
    \begin{subfigure}{\textwidth}
        \centering
        \includegraphics[width=0.49\linewidth]{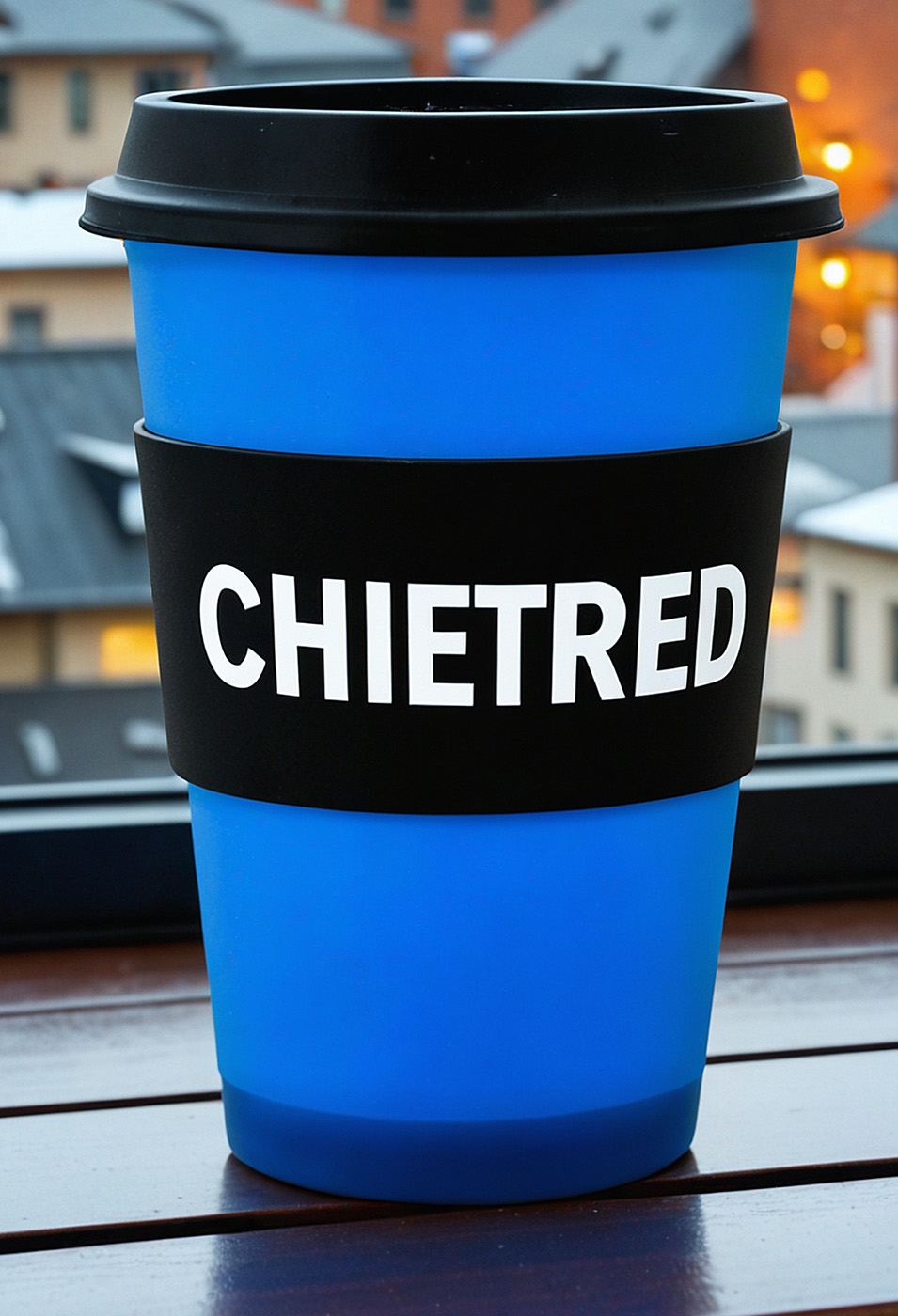}\hfill
        \includegraphics[width=0.49\linewidth]{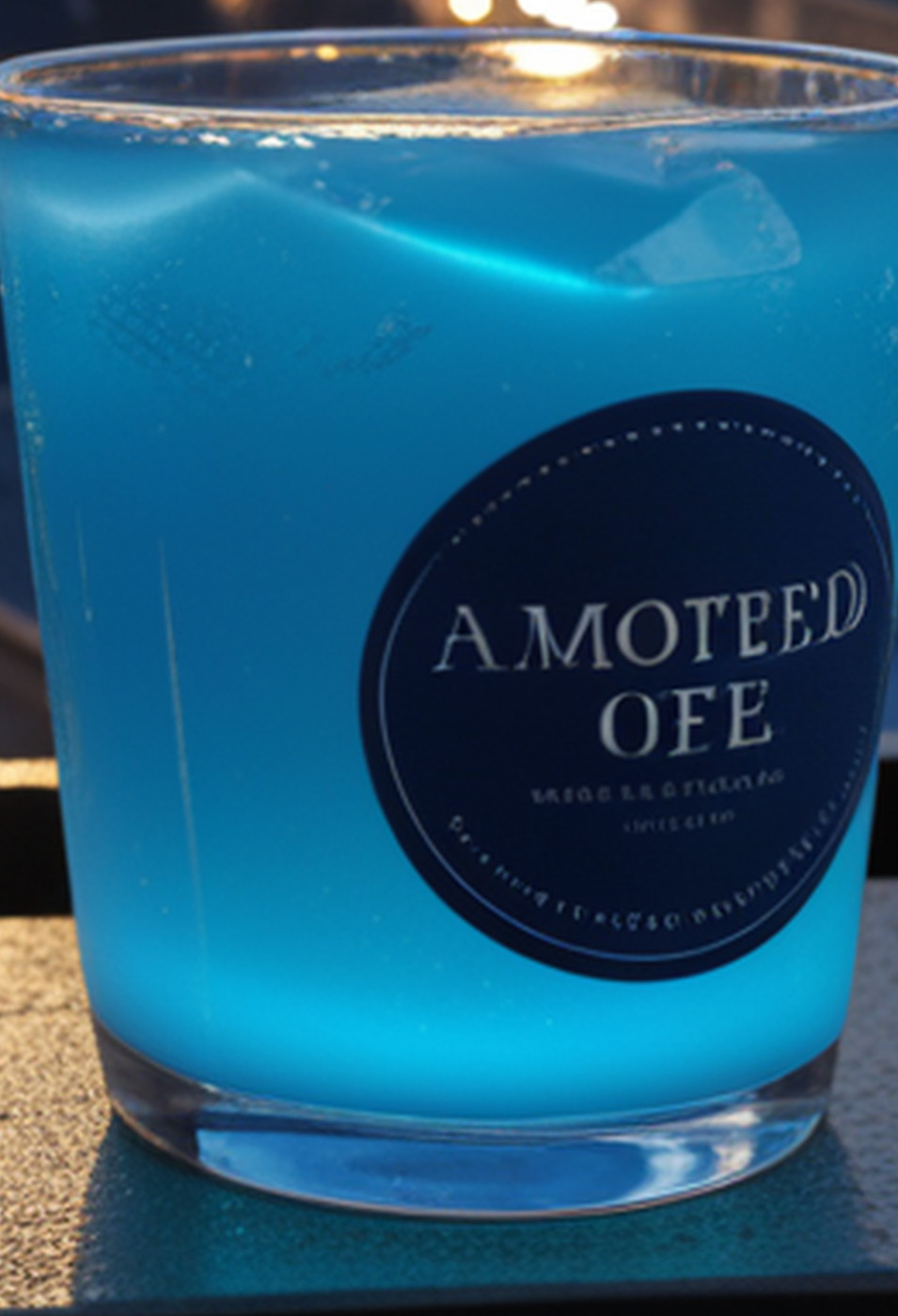}
        \caption{Cup, zoom-in: sharper edges and bokeh, fewer artifacts; legible letters.}
        \label{fig:cup_qual_crop}
    \end{subfigure}
    \caption{Qualitative comparison (cup). Left: CADE~2.5 with QSilk. Right: baseline. Same prompts and seed.}
    \label{fig:cup_qual}
\end{figure}

\paragraph{Setup.} Models: finetuned SDXL (illustrious); resolutions: 5K long side; steps: 20--50; hardware: single 24--48\,GB GPU; attention accel optional.
\paragraph{Baselines.} Stock (no clamp), global hard/soft clip.
\paragraph{Evaluation.} We rely on qualitative, side-by-side inspection with fixed seeds and prompts (Figs.~\ref{fig:portrait_qual}, \ref{fig:dog_qual}, \ref{fig:cup_qual}). Across cases, QSilk yields cleaner high-frequency detail, fewer halos/moire, and more coherent small structure; on SDXL, letterforms are noticeably more legible (zoom in Fig.~\ref{fig:cup_qual_crop}). A thorough quantitative study is left to future work.

\medskip
\noindent\textbf{Repro setup used in figures.} Unless noted otherwise, we use finetuned SDXL (illustrious) backbones with CADE~2.5 integration and QSilk enabled. Common parameters: \texttt{seed=23132}, \texttt{steps=30}, \texttt{CFG=7}, \texttt{sampler=DDIM}, \texttt{denoise=1.0}. Prompts follow three themes: photoportrait, white dog, and cup of coffee (see captions).

\FloatBarrier

\FloatBarrier

\FloatBarrier

\noindent\textbf{Observation on text rendering.} Beyond generally cleaner micro-texture and more stable bokeh, we find that fine semantic details become more \emph{coherent, correct, and crisp}. A notable side effect on SDXL is \textbf{stabilized letterforms}: generated text becomes more legible with QSilk in CADE~2.5 (see Fig.~\ref{fig:cup_qual}, zoom in Fig.~\ref{fig:cup_qual_crop}), suggesting that tail suppression benefits character-level structure as well.

\section{Related Work}
DDPM~\cite{ho2020ddpm}, LDM~\cite{rombach2022ldm}, and CFG~\cite{ho2021cfg} underpin modern T2I systems. Imagen's dynamic thresholding~\cite{saharia2022imagen, saharia2022imagenSup} clips predicted $x_0$ percentiles in \emph{image space}; public implementations caution against applying it in latent space~\cite{diffusersDynamicThresh}. S-CFG~\cite{shen2024scfg} adjusts \emph{guidance} spatially. Recent works adapt CFG over time~\cite{malarz2025betaCfg}. Uncertainty guidance via entropy/margin has been explored for diffusion sampling~\cite{luo2024measurementGuidance}. Our method is complementary: we perform \emph{local, confidence-aware amplitude regularization} in latent space with seam-free tiling.

\section{Limitations and Ethical Considerations}
When attention maps are unavailable, the Lite proxy helps but may under-adapt on extremely flat regions. Very aggressive softness $\alpha$ or tight quantiles can under-expose fine texture. AQClip assumes stationary statistics within a tile; very thin structures may benefit from smaller stride. QSilk improves visual fidelity without enabling misuse beyond existing diffusion capabilities; we recommend transparent disclosure when images are enhanced.

\section{Reproducibility}
Code, presets, and exact seeds will be released upon publication. We include reference implementations for SD/SDXL and CADE~2.5 integration.

\section{Conclusion}
QSilk offers a principled, training-free way to stabilize latent diffusion while preserving micro-texture. Its adaptive quantile corridor---optionally guided by attention entropy---yields cleaner, sharper results with negligible overhead and integrates seamlessly with CADE~2.5.

\section*{Acknowledgments}
The author used GPT-5 to assist with drafting, editing, code suggestions, and figure layout. All technical decisions, implementations, experiments, and validation were performed by the human author, who takes full responsibility for the content.

{\small
\bibliographystyle{plain}
\bibliography{refs_qsilk}
}

\appendix
\section{Implementation Notes}
\begin{itemize}
\item Seamless overlap via fold weight normalization prevents seams at tile borders.
\item Attention-entropy probe: sub-sample heads and tokens to bound runtime; normalize per sample.
\item Compute stats in FP32 for stability; cast back to original dtype.
\item Defaults chosen for robustness across SD/SDXL; per-project tuning rarely needed.
\end{itemize}

\end{document}